\pdfoutput=1
\PassOptionsToPackage{table,dvipsnames}{xcolor}  

\documentclass[11pt]{article}

\usepackage{amsmath}
\usepackage{titling}

\usepackage[final]{acl}
\usepackage{booktabs}
\usepackage{hyperref}
\usepackage[utf8]{inputenc}
\usepackage{xcolor}
\usepackage[normalem]{ulem}

\definecolor{PrimaryBlue}{RGB}{0,0,255}
\definecolor{PrimaryRed}{RGB}{255,0,0}

\definecolor{TaskBG}{RGB}{235,240,255}

\DeclareRobustCommand{\cmark}{\textcolor{PrimaryBlue}{\ding{51}}} 
\DeclareRobustCommand{\xmark}{\textcolor{PrimaryRed}{\ding{55}}}  
\definecolor{LightGray}{gray}{0.92}

\definecolor{RowBG}{gray}{0.96}
\definecolor{BlueAdd}{RGB}{0,0,200}
\definecolor{RedDel}{RGB}{180,0,0}

\usepackage{times}
\usepackage{latexsym}
\usepackage{multirow}
\usepackage{tabularx}
\usepackage{pifont}
\usepackage{tcolorbox}
\usepackage{xcolor}
\usepackage{multirow}      
\usepackage{booktabs}      
\usepackage{adjustbox}     
\usepackage{caption}       
\usepackage{geometry}      
\usepackage{amssymb}
\usepackage[T1]{fontenc}
\usepackage[utf8]{inputenc}
\usepackage{textcomp}
\usepackage{microtype}
\usepackage{inconsolata}
\usepackage{seqsplit}         

\usepackage{xcolor}
\usepackage{graphicx}
\usepackage{framed}
\usepackage{pifont}
\definecolor{shadecolor}{rgb}{0.92,0.92,0.92}
\usepackage{tikz}
\usetikzlibrary{shapes.geometric, arrows.meta, positioning}

\tikzstyle{box} = [rectangle, rounded corners, minimum width=3.2cm, minimum height=1cm, text centered, draw=black, fill=gray!10]
\tikzstyle{process} = [rectangle, minimum width=3.2cm, minimum height=1cm, text centered, draw=black, fill=blue!10]
\tikzstyle{decision} = [diamond, draw=black, fill=yellow!30, minimum size=1.2cm, text centered, inner sep=0pt, aspect=2]
\tikzstyle{arrow} = [thick, ->, >=stealth]
\PassOptionsToPackage{table}{xcolor}

\title{Bridging the Editing Gap in LLMs: FineEdit for Precise and Targeted Text Modifications$^{\ddag}$}

\author{%
\textbf{Yiming Zeng}$^{1}$\thanks{Equal contribution.}, 
\textbf{Wanhao Yu}$^{2}$\footnotemark[1], 
\textbf{Zexin Li}$^{3}$, 
\textbf{Tao Ren}$^{4}$,\\
\textbf{Yu Ma}$^{5}$, 
\textbf{Jinghan Cao}$^{6}$, 
\textbf{Xiyan Chen}$^{4}$, 
\textbf{Tingting Yu}$^{1}$\thanks{Corresponding author.} \\[2mm]
$^{1}$University of Connecticut,\;
$^{2}$University of North Carolina at Charlotte,\\
$^{3}$University of California, Riverside,\;
$^{4}$University of Pittsburgh,\\
$^{5}$Carnegie Mellon University,\;
$^{6}$San Francisco State University \\[2mm]
$^{1}$\{yiming.zeng, tingting.yu\}@uconn.edu,\;
$^{2}$wyu6@charlotte.edu,\;
$^{3}$zli536@ucr.edu,\\
$^{4}$\{tar118, xic130\}@pitt.edu,\;
$^{5}$yuma13926@gmail.com,\;
$^{6}$jcao3@alumni.sfsu.edu \\
}

\usepackage{float}
\begin{document}

\maketitle
\begingroup
\renewcommand\thefootnote{\ddag}
\footnotetext{This work has been accepted to EMNLP 2025.}
\endgroup
\vspace{10em}
\begin{abstract}

Large Language Models (LLMs) have significantly advanced natural language processing, demonstrating strong capabilities in tasks such as text generation, summarization, and reasoning. Recently, their potential for automating precise text editing tasks across specialized domains, such as programming code, LaTeX, and structured database languages, has gained attention. However, current state-of-the-art LLMs still struggle with executing precise, instruction-driven edits, particularly when structural accuracy and strict adherence to domain conventions are required.
To address these challenges, we introduce InstrEditBench, an automated benchmark dataset comprising over 30,000 structured editing tasks spanning diverse domains, including Wikipedia articles, LaTeX documents, source code, and database languages. Using this benchmark, we develop FineEdit, a specialized editing model explicitly trained for accurate, context-aware text modifications. Experimental evaluations demonstrate that FineEdit outperforms state-of-the-art models, achieving improvements of approximately 10\% over Gemini models on single-turn edits, up to 30\% over Llama-3.2-3B, and exceeding Mistral-7B-OpenOrca performance by over 40\% on direct editing tasks. FineEdit also effectively generalizes to realistic multi-turn editing scenarios, highlighting its practical applicability. To facilitate further research and reproducibility, we release FineEdit at \url{https://github.com/StuRinDQB/FineEdit} and \url{https://huggingface.co/datasets/YimingZeng/FineEdit_bench}.

\end{abstract}

\section{Introduction}
Large Language Models (LLMs) have brought transformative progress to the field of natural language processing, demonstrating remarkable capabilities in text generation, summarization, and reasoning~\cite{chen-etal-2022-generate, achiam2023gpt,chen-etal-2023-dynamic,chen-etal-2024-beyond-single, regulogpt, li-etal-2025-learning-committee, liang2025autoranweaktostrongjailbreakinglarge}.
Recently, LLMs have received increasing attention for their potential to automate and enhance text editing across a variety of domains~\cite{celikyilmaz2020evaluation}. Such editing capabilities is particularly needed under task-specific application scenarios, e.g., code editing~\cite{fan2024exploringcapabilitiesllmscode, lei2025infantagentnextmultimodalgeneralistagent}, Wiki editing~\cite{suri2024docedit}, etc. 

Despite this promise, current LLMs still face notable limitations when applied to tasks that demand direct editing, where the model must simultaneously understand the original text, follow the instruction precisely, and generate semantically aligned, high-quality edits. Even powerful proprietary tools like ChatGPT often struggle to fully understand user intent and reliably follow strict editing instructions, especially in long-context scenarios~\cite{castillo2022chat}. Particularly, LLMs’ general editing capabilities in task-specific settings often fall short~\cite{yao-etal-2023-editing, ma-etal-2024-robustness}. 
They tend to generate incorrect outputs and stray from the given editing instructions. 

To address these challenges, we propose a more focused approach to editing with LLMs. Our key insight is that narrowing the model’s attention to two fundamental aspects, the exact location of the edit and the content to be modified, can significantly improve performance in direct editing tasks. Per this intuition, we propose a dual approach consisting of a dedicated benchmark (InstrEditBench) for editing tasks and an editing-specific model (FineEdit). Specifically, we design an automated workflow that focuses on accurately identifying and evaluating structured text edits. This workflow identifies precise differences and ensures correct edits through quality control. By reducing noise and focusing on meaningful modifications, this process produces a dedicated, high-quality benchmark. It directly addresses limitations in existing methods and aligns better with the practical demands of real-world editing tasks. Notably, our approach is also generalized to multi-turn editing scenarios, a much more realistic user scenario, where instructions arrive iteratively and allow the model to refine its edits step by step. 

\noindent \textbf{Implementation and evaluation.} We train the FineEdit model on InstrEditBench benchmark, explicitly designed to optimize performance on direct, instruction-driven text editing tasks. The result shows that FineEdit achieves an improvement of 10\% over Gemini 1.5 Flash and Gemini 2.0 Flash~\cite{google2024gemini} in single-turn editing tasks, and up to 30\% over Llama-3.2-3B~\cite{meta2024llama3_2} on diverse editing benchmarks, while outperforming Mistral-7B-OpenOrca~\cite{lian2023mistralorca1, mukherjee2023orca, longpre2023flan} over 40\% on direct editing tasks. 

The main contributions of this work include:
\begin{itemize}
    \item \textbf{A high-quality benchmark (InstrEditBench)}: We introduce the first systematically constructed benchmark that spans four diverse domains and contains more than {30,000} single-turn and multi-turn structured editing tasks, thereby establishing a unified and comprehensive evaluation standard for direct editing research.
    
    \item \textbf{An innovative automated dataset generation workflow}: We develop a comprehensive workflow that ensures the benchmark's quality by accurately identifying line numbers and applying rigorous criteria to filter meaningful and relevant edits.
    
    \item \textbf{The FineEdit model}: We present a specialized model designed for direct text editing, demonstrating superior performance across benchmarks compared with existing models.
\end{itemize}

\section{Background}
\subsection{Problem Formulation}
Each data point consists of an original structured text, \(T_{\text{orig}}\), and an editing instruction, \(I_{\text{edit}}\). The objective is to generate an edited text, \(T_{\text{edit}}\), that incorporates the modifications specified by \(I_{\text{edit}}\). Formally, this process is defined as
\begin{equation}
    T_{\text{edit}} = f\Bigl(T_{\text{orig}}, I_{\text{edit}}; \theta\Bigr)
\end{equation}
where \(\theta\) represents learned parameters and \(f\) denotes a function instantiated by a LLM that maps the original text \(T_{\text{orig}}\) and editing instruction \(I_{\text{edit}}\) to the edited text \(T_{\text{edit}}\).

The parameters \(\theta\) are learned from a dataset consisting of triples \(\{(T_{\text{orig}}^{(i)}, I_{\text{edit}}^{(i)}, T_{\text{edit}}^{(i)})\}_{i=1}^{N}\) during training, where the objective is to minimize the discrepancy between the generated output and the ground truth edited text.

Internally, \(f\) concatenates \(T_{\text{orig}}\) and \(I_{\text{edit}}\) into a single prompt and generates \(T_{\text{edit}}\) token by token in an autoregressive manner. Specifically, if 
\(T_{\text{edit}} = (y_1, y_2, \dots, y_t)\),
the probability of the edited text is factorized as
\begin{equation}
\begin{split}
        p(T_{\text{edit}} \mid T_{\text{orig}}, I_{\text{edit}}) =& \prod_{i=1}^{t} p\Bigl(y_i \mid T_{\text{orig}}, I_{\text{edit}}, \\
        & y_1, y_2, \dots, y_{i-1}\Bigr)
\end{split}
\end{equation}

For finetuning on the editing task, the prompt tokens (i.e., the original text and the editing instruction) are masked out in the loss function to ensure that the model focuses only on predicting the correct edited tokens. At inference time, the model processes the prompt and subsequently generates \(T_{\text{edit}}\).

The parameters \(\theta\) are fine-tuned on labeled examples \((T_{\text{orig}}, I_{\text{edit}}, T_{\text{edit}})\) by minimizing the negative log-likelihood of the target tokens with the loss:
\begin{equation}
    \mathcal{L}(\theta) = - \sum_{t=1}^{|T_{\text{edit}}|} \log P_{\theta}(y_t \mid T_{\text{orig}}, I_{\text{edit}}, y_{1:t-1})
\end{equation}
over all training samples in the dataset.

\subsection{LLM Editing Tasks}

LLMs are increasingly recognized as versatile tools for automating and enhancing editing tasks across diverse domains. Previous studies have explored LLMs for editing tasks in areas such as natural language (e.g., wiki articles) and code. For instance, CoEdIT~\cite{raheja2023coedit} employs task-specific instruction tuning to achieve precise modifications, while other works fine-tune models like T5~\cite{raffel2020exploring} on pairs of original and edited texts~\cite{ bryant2019bea, stahlberg2021synthetic, pezeshkpour2023measuring}. However, many of these approaches rely on specialized techniques or focus narrowly on specific tasks, such as grammar correction~\cite{katinskaia2023grammatical, bout2023efficient}, text simplification~\cite{sun2023teaching}, paraphrase generation~\cite{palivela2021optimization}, or style transfer~\cite{luo2023prompt}, which limits their generalizability across a broader range of editing scenarios. In the realm of code editing, Dilhara et al.~\cite{dilhara2024unprecedented} examined LLMs for code change tasks and identified weaknesses in generating accurate reviews and commit messages.  
Beyond single-turn editing, iterative or multi-turn editing can further improve output quality by allowing incorporation of progressive feedback, leading to more accurate and context-aligned modifications~\cite{schick2022peer, madaan2023self}. 
While these studies offer valuable insights, they often fall short in providing unified benchmarks and robust solutions to address the full spectrum of editing challenges. Our work addresses these gaps by introducing a comprehensive, cross-scenario editing tasks benchmark that covers Wiki, code, DSL, and LaTeX.

\begin{figure*}[!htbp]
    \centering
    \includegraphics[width=0.8\textwidth]{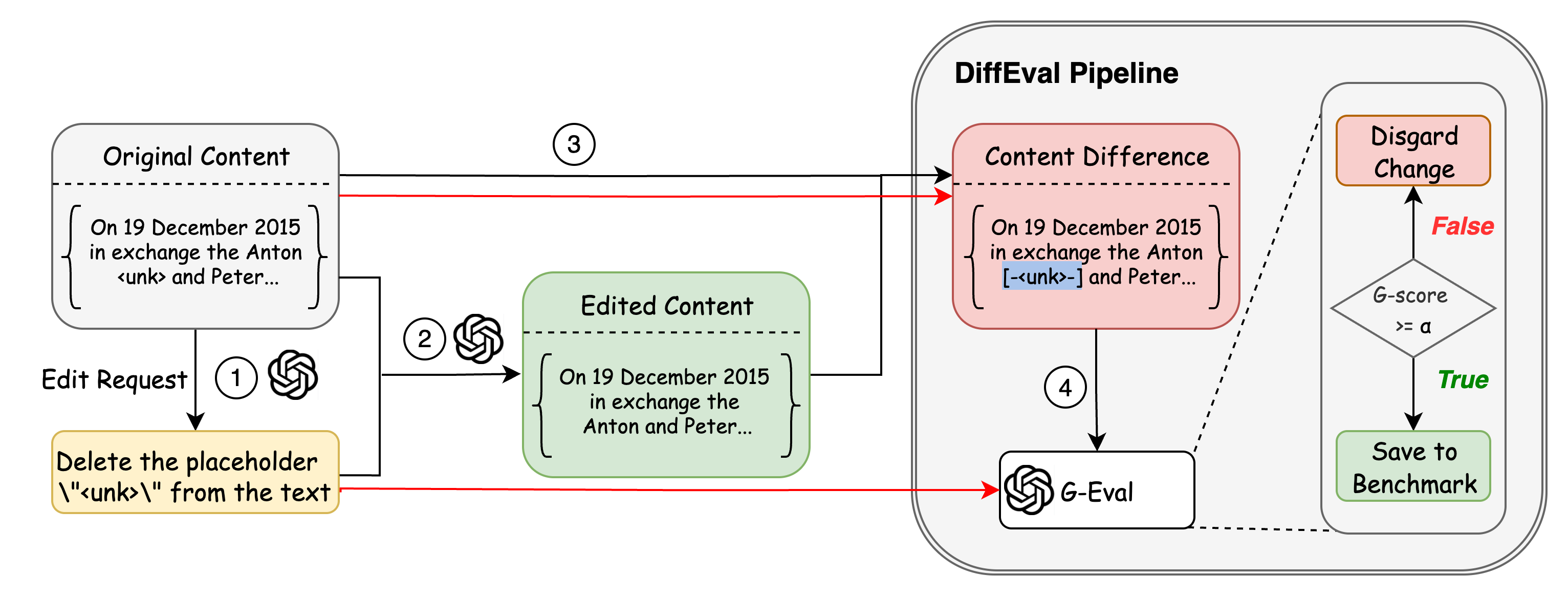}
    \caption{Workflow of Generating High-quality InstrEditBench. The content difference is highlighted in blue. }
    \label{fig:example-pdf}
    \vspace{-5mm}
\end{figure*}

\section{Method}

\subsection{Instruction categories}

We leverage four data sources to cover a wide range of representative text application scenarios: Wiki, Code, DSL, and LaTeX. The details of each categories are described as follows:

\begin{itemize}
    \item \textbf{Wiki}: Data is extracted from the WikiText language modeling dataset~\cite{merity2016pointer}, which contains over 100 million tokens from a dedicated subset of Wikipedia's Good articles~\cite{wikipedia_good_articles} and Wikipedia's Featured articles~\cite{wikipedia_featured_articles}. Specifically, sections from these articles are extracted and then contiguous segments are randomly selected to provide data points with various lengths.
    \item \textbf{Code}: Code samples are extracted from the CodeSearchNet corpus \cite{husain2019codesearchnet}, which contains about two million pairs of comments and code from GitHub projects. To make the edit task more challenging, each code sample in our benchmark is made up of several instead of one code segment because one single code segment is too short (about 10 lines).
    \item \textbf{DSL}: Database Domain Specific Language (DSL) is also considered in our benchmark. It consists of queries and schema definitions from multiple public repositories~\cite{b-mc2_2023_sql-create-context, hive,cassandra,chinookDatabase}.
    \item \textbf{LaTeX}: LaTeX data is extracted from the Latex2Poster dataset~\cite{latex2poster} that offers the LaTeX source code document of research papers along with metadata. Specifically, each data point in our benchmark consists of multiple subsections from each extracted document data.
\end{itemize}

\subsection{Instruction Generation}

Zero-shot instruction generation is efficient, but often lacks diversity. To address this limitation, we build on the work of \cite{wang2022self, alpaca} by leveraging ChatGPT-4o mini combined with in-context learning (ICL)~\cite{dong2024survey}. Our approach is designed to generate specific edit requests tailored to the structural characteristics of different data categories, as process \ding{192} in Figure~\ref{fig:example-pdf}. For Wiki, which primarily consists of clear structural text elements like headings and subheadings, we apply a zero-shot prompting strategy. In contrast, for more complex domains such as LaTeX, code, and DSL, we adopt ICL to improve the diversity and nuance of generated instructions. 

This category-specific strategy not only enriches the instruction sets but also enhances their ability to capture domain-specific editing challenges without compromising on precision and efficiency. We will describe prompt details in Appendix~\ref{sec:dataset_generation_prompts}.


\subsection{Instruction filtering}

After obtaining the edit instructions for each content, we apply them to the original text to produce an edited version as process \ding{193} in Figure~\ref{fig:example-pdf}. However, ensuring the quality of the edited content remains challenging. Although LLM generally follows the edit instructions, errors may occur---for example, targeting incorrect line numbers or misinterpreting the intended semantics~\cite{cassano2024editevaluatingabilitylarge, wang2025understandingcharacteristicscodegeneration}. To address this problem and improve data quality, we propose {DiffEval Pipeline}, which integrates G-Eval~\cite{liu2023geval} and Git-Diff as an automatic filter to improve data quality. 


Besides adopting G-Eval for automated assessment \citep{liu2023geval}, the DiffEval Pipeline also relies on \texttt{git} \citep{gitdiff}, a widely used version control system, to detect and classify textual modifications. Specifically, the command \texttt{git diff} specifies differences between the original and modified texts as process \ding{194} in Figure~\ref{fig:example-pdf}, categorizing changes into four types:

\begin{itemize}
\item \textbf{Replacements}: an original segment is transformed into a new form, indicated as \texttt{[original\_text -> modified\_text]}. This captures cases where an existing text portion is substituted with different content, which may alter meaning or style.
\item \textbf{Deletions}: a segment is removed entirely, shown as \texttt{[-original\_text-]}. Such removals can simplify the text or eliminate irrelevant or erroneous sections.
\item \textbf{Insertions}: new content is added, denoted as \texttt{[+modified\_text+]}. Insertions enrich the text with extra details, clarifications, or elaborations.
\item \textbf{Unchanged Text}: labeled as \texttt{equal: unchanged\_text}. This indicates portions that remain identical between the original and modified versions, providing a reference for what the model has chosen to retain.
\end{itemize}

By categorizing changes into these four types, the DiffEval Pipeline offers a structured view of how text is altered, enabling more precise evaluations when paired with G-Eval.

Finally, process \ding{195} in Figure~\ref{fig:example-pdf} demonstrates that DiffEval carefully reviews the aggregated data (marked with red arrows) alongside the edit request to fully grasp the context, structure, and nuances of the text. It identifies discrepancies between the intended edits and the actual modifications, verifying whether the changes faithfully implement the edit instructions. By using the \texttt{git diff} output instead of the complete edited content, DiffEval can precisely locate modifications using supplementary information such as line numbers and structured differences. Moreover, \texttt{git diff} minimizes unnecessary noise and reduces computational overhead by significantly lowering the token count compared with the full edited content. Once all required data is gathered, the G-Eval analysis process evaluates the collected information to further enhance the dataset quality.

Specifically, the analysis process begins by parsing the structure of \texttt{git diff} outputs, categorizing changes as replacements, deletions, insertions, or unchanged segments. Next, it evaluates the semantic meaning of both the original content and the modifications to ensure that the changes are accurate and complete. This involves a thorough review of the original text, the edit request, and the resulting edits, applying predefined categorization rules, and assessing overall coherence.

Based on this analysis process, the DiffEval can assign a coherence score, \texttt{G-Score}, to the edited content, reflecting the semantic integrity and logical consistency of the modifications. This score is used to filter out output that does not meet the desired quality threshold $\alpha$.


\subsection{Generalize to Multi-turn Editing Task}
Notably, our proposed framework is easily generalized to more practical multi-turn editing scenarios. Specifically, given the initial content, we instruct ChatGPT to generate a sequence of multiple distinct editing requests that are explicitly constrained to be non-contradictory with each other. Each generated editing request targets different aspects or details within the same content, ensuring that subsequent instructions complement rather than conflict with previous edits. This setting reflects real-world editing workflows where users iteratively refine content through consecutive instructions.

\section{Experiment}
\subsection{Experimental Setup}
In this section, we detail the experimental setups, including dataset splits, model variants, baselines, evaluation metrics, and implementation specifics.

\noindent \textbf{Dataset and Model Variants.} We evaluate FineEdit on our proposed InstrEditBench dataset using a 90/10 train-test split. Additionally, we introduce three FineEdit variants, namely FineEdit-L, FineEdit-XL, and FineEdit-Pro, which are fine-tuned from the LLaMA-3.2-1B, LLaMA-3.2-3B, and Qwen2.5-3B-Instruct base models respectively, covering a broad spectrum of model architectures and parameter scales.

\noindent \textbf{Baselines.} Our baselines include Gemini 1.5 Flash, Gemini 2.0 Flash, LLaMA-3.2-1B, LLaMA-3.2-3B, Qwen2.5-3B-Instruct, and Mistral-7B, spanning diverse architectures and sizes. We evaluate both zero-shot and few-shot prompting on the Gemini models, while open-source models are assessed using zero-shot prompting.

\noindent \textbf{Metrics.} Following established approaches~\cite{shen2017style, nakamachi2020text}, we use BLEU and ROUGE‑L metrics to assess the vocabulary and structural consistency between the edited and reference texts.

\noindent \textbf{Implementation details.} 
Training details are provided in Appendix~\ref{sec:appx_implementation_detail}.
\begin{table*}[!tbp]
    \centering
    \small
    \renewcommand{\arraystretch}{1.35}
    \begin{adjustbox}{width=\textwidth}
    \begin{tabular}{llc|cc|cc|cc|cc|cc}
        \toprule
        \multirow{2}{*}{\textbf{Method}} &
        \multirow{2}{*}{\textbf{Model}} &
        \multirow{2}{*}{\textbf{Open-Source}} &
        \multicolumn{2}{c|}{\textbf{LaTeX}} &
        \multicolumn{2}{c|}{\textbf{DSL}} &
        \multicolumn{2}{c|}{\textbf{Wiki}} &
        \multicolumn{2}{c|}{\textbf{Code}} &
        \multicolumn{2}{c}{\textbf{Overall}} \\
        \cmidrule(lr){4-5}\cmidrule(lr){6-7}\cmidrule(lr){8-9}\cmidrule(lr){10-11}\cmidrule(l){12-13}
        &  &  &
        \textbf{BLEU} & \textbf{RG-L} &
        \textbf{BLEU} & \textbf{RG-L} &
        \textbf{BLEU} & \textbf{RG-L} &
        \textbf{BLEU} & \textbf{RG-L} &
        \textbf{BLEU} & \textbf{RG-L} \\
        \midrule
\multirow{6}{*}{Zero-shot}
        & Gemini 1.5 Flash  & \ding{55} & 0.8665 & 0.9150 & 0.8297 & 0.8555 & 0.7626 & 0.8361 & 0.8551 & 0.9073 & 0.8285 & 0.8819 \\
        & Gemini 2.0 Flash  & \ding{55} & 0.7413 & 0.7951 & 0.4706 & 0.4964 & 0.9133 & 0.9429 & 0.1339 & 0.2737 & 0.5853 & 0.6519 \\
        & Llama-3.2-1B      & \checkmark & 0.5088 & 0.6108 & 0.5564 & 0.6596 & 0.4413 & 0.5766 & 0.4742 & 0.6072 & 0.4867 & 0.6069 \\
        & Llama-3.2-3B      & \checkmark & 0.5969 & 0.6925 & 0.5747 & 0.6821 & 0.5061 & 0.6384 & 0.6638 & 0.7727 & 0.5862 & 0.6976 \\
        & Qwen2.5-3B-Instr  & \checkmark & 0.5467 & 0.6712 & 0.4107 & 0.4991 & 0.4170 & 0.5699 & 0.3967 & 0.5390 & 0.4492 & 0.5816 \\
        & Mistral-7B-Orca   & \checkmark & 0.3782 & 0.5770 & 0.0361 & 0.1638 & 0.3608 & 0.5840 & 0.3763 & 0.6447 & 0.3246 & 0.5395 \\
\cmidrule{1-13}
\multirow{2}{*}{Few-shot}
        & Gemini 1.5 Flash (2 shot) & \ding{55} & 0.8742 & 0.9324 & 0.0908 & 0.1190 & 0.8657 & 0.9139 & 0.7412 & 0.8302 & 0.7249 & 0.7845 \\
        & Gemini 2.0 Flash (2 shot) & \ding{55} & 0.9464 & 0.9723 & 0.1600 & 0.1814 & \textbf{0.9380} & \textbf{0.9665} & 0.8327 & 0.8698 & 0.8011 & 0.8302 \\
\cmidrule{1-13}
\multirow{3}{*}{FineEdit}
        & FineEdit-L   & \checkmark & 0.9311 & 0.9697 & 0.9334 & 0.9615 & 0.8077 & 0.9036 & 0.9296 & 0.9725 & 0.8957 & 0.9504 \\
        & FineEdit-XL  & \checkmark & 0.8867 & 0.9502 & 0.9241 & 0.9552 & 0.8120 & 0.9056 & 0.9295 & 0.9720 & 0.8824 & 0.9441 \\
        & FineEdit-Pro & \checkmark & \textbf{0.9539} & \textbf{0.9821} & \textbf{0.9521} & \textbf{0.9710} & 0.8521 & 0.9185 & \textbf{0.9538} & \textbf{0.9836} & \textbf{0.9245} & \textbf{0.9628} \\
        \bottomrule
    \end{tabular}
    \end{adjustbox}
    \caption{Comparison of LLMs on BLEU and ROUGE-L for LaTeX, DSL, Wiki, Code. Overall data displays
average performance among all data categories. The best results are highlighted in bold.}
    \label{tab:llm_single}
    \vspace{-3mm}
\end{table*}

\begin{table*}[!t]
    \centering
    \small
    \renewcommand{\arraystretch}{1.35}
    \begin{adjustbox}{width=\textwidth}
    \begin{tabular}{l l c | cc|cc|cc|cc|cc}
        \toprule
        \multirow{2}{*}{\textbf{Method}} &
        \multirow{2}{*}{\textbf{Model}} &
        \multirow{2}{*}{\textbf{Open-Source}} &
        \multicolumn{2}{c|}{\textbf{LaTeX}} &
        \multicolumn{2}{c|}{\textbf{DSL}} &
        \multicolumn{2}{c|}{\textbf{Wiki}} &
        \multicolumn{2}{c|}{\textbf{Code}} &
        \multicolumn{2}{c}{\textbf{Overall}} \\
        \cmidrule(lr){4-5}\cmidrule(lr){6-7}\cmidrule(lr){8-9}\cmidrule(lr){10-11}\cmidrule(l){12-13}
        &  &  &
        \textbf{BLEU} & \textbf{RG-L} &
        \textbf{BLEU} & \textbf{RG-L} &
        \textbf{BLEU} & \textbf{RG-L} &
        \textbf{BLEU} & \textbf{RG-L} &
        \textbf{BLEU} & \textbf{RG-L} \\
        \midrule
\multirow{5}{*}{Zero-shot}
        & Gemini 1.5 Flash  & \ding{55} & 0.1745 & 0.3067 & 0.9643 & 0.9787 & 0.7785 & 0.8908 & 0.3882 & 0.5291 & 0.5764 & 0.6763 \\
        & Gemini 2.0 Flash  & \ding{55} & 0.1304 & 0.2435 & 0.4243 & 0.4328 & 0.8624 & 0.9145 & 0.1426 & 0.2454 & 0.3899 & 0.4591 \\
        & Llama-3.2-1B      & \checkmark & 0.3168 & 0.4371 & 0.2593 & 0.3466 & 0.2291 & 0.3584 & 0.1749 & 0.3196 & 0.2450 & 0.3654 \\
        & Llama-3.2-3B      & \checkmark & 0.3101 & 0.4374 & 0.3274 & 0.4355 & 0.2871 & 0.4073 & 0.2988 & 0.4177 & 0.3059 & 0.4245 \\
        & Qwen2.5-3B-Instr  & \checkmark & 0.5196 & 0.6344 & 0.2083 & 0.2603 & 0.2845 & 0.4261 & 0.3985 & 0.5138 & 0.3527 & 0.4587 \\
\cmidrule{1-13}
\multirow{2}{*}{Few-shot}
        & Gemini 1.5 Flash (2 shot) & \ding{55} & 0.4811 & 0.5423 & 0.0511 & 0.1167 & 0.7511 & 0.8462 & 0.2388 & 0.3430 & 0.3805 & 0.4621 \\
        & Gemini 2.0 Flash (2 shot) & \ding{55} & 0.9099 & 0.9247 & 0.0294 & 0.0406 & 0.9272 & 0.9740 & 0.4719 & 0.6266 & 0.5846 & 0.6415 \\
\cmidrule{1-13}
\multirow{3}{*}{FineEdit}
        & FineEdit-L   & \checkmark & 0.6823 & 0.8531 & 0.8071 & 0.8730 & 0.4938 & 0.6588 & 0.6707 & 0.7773 & 0.6635 & 0.7906 \\
        & FineEdit-XL  & \checkmark & 0.3230 & 0.4468 & 0.8050 & 0.8798 & 0.4522 &  0.6333 & 0.6806 & 0.7756 & 0.5652 & 0.6839 \\
        & FineEdit-Pro & \checkmark & 0.8461 & 0.8917 & 0.8123 & 0.8902 & 0.6975 & 0.8286 & 0.9499 & 0.9796 & 0.8265 & 0.8975 \\
        \bottomrule
    \end{tabular}
    \end{adjustbox}
    \caption{Multi-turn editing results for LaTeX, DSL, Wiki and Code. Overall data displays
average performance among all data categories.}
    \label{tab:llm_multi}
    \vspace{-5mm}
\end{table*}

\subsection{Performance of Existing Models}

We evaluated FineEdit against several state-of-the-art baselines on the InstrEditBench dataset across four data categories as presented in Table~\ref{tab:gemini_finedit_comparison}.

\noindent \textbf{Comparison with Zero-shot Performance.} Among all baselines, Gemini 1.5 Flash achieved the highest overall scores, while Mistral-7B-OpenOrca recorded the lowest BLEU and ROUGE-L values. Although model size is typically a critical factor, Gemini 2.0 Flash did not outperform Gemini 1.5 Flash in terms of overall effectiveness. For example, despite having more parameters than LLaMA-3.2-1B, Mistral-7B-OpenOrca underperformed on both metrics, highlighting the importance of model architecture and training strategies. Additionally, Gemini 2.0 Flash demonstrated superior semantic understanding in the Wiki category, with a BLEU score of 0.9133 and a ROUGE-L score of 0.9429, yet its overall performance remained inferior to that of Gemini 1.5 Flash.

FineEdit, and in particular its FineEdit-Pro variant, further outperforms all zero-shot baselines. FineEdit-Pro achieves an overall BLEU score of 0.9245, representing improvements of approximately 11.6\%, 57.7\%, and 184.7\% over Gemini 1.5 Flash (0.8285), LLaMA-3.2-3B (0.5862), and Mistral-7B-OpenOrca (0.3246), respectively. These gains are consistently observed across individual data categories—for example, FineEdit-Pro attains BLEU scores of 0.9521 and 0.9538 in the DSL and Code domains, respectively. These results underscore the effectiveness of FineEdit’s targeted fine-tuning strategy, which focuses on precise editing of location and content to preserve both structural and semantic integrity.

\noindent \textbf{Comparison with Few-shot Performance.} We further evaluated few-shot learning on the Gemini models. Although few-shot prompting notably improved performance in some categories, such as the LaTeX domain where Gemini 2.0 Flash achieved a 20\% higher BLEU score compared to the zero-shot setting, the overall few-shot results still remained inferior to FineEdit. In certain cases, such as the SQL category, few-shot learning provided minimal improvement, achieving BLEU and ROUGE-L scores of only 0.1600 and 0.1814, respectively. These findings highlight the effectiveness and importance of our curated benchmark in driving advancements in editing tasks.

\definecolor{PrimaryBlue}{RGB}{0,0,255}
\definecolor{PrimaryRed}{RGB}{255,0,0}

\definecolor{TaskBG}{RGB}{235,240,255}

\DeclareRobustCommand{\cmark}{\textcolor{PrimaryBlue}{\ding{51}}} 
\DeclareRobustCommand{\xmark}{\textcolor{PrimaryRed}{\ding{55}}}  

\subsection{FineEdit: Supervised Finetuning} 
Our FineEdit model is offered in three variants: FineEdit-L, FineEdit-XL, and FineEdit-Pro. Under zero-shot conditions, FineEdit-L consistently outperforms all baseline models in BLEU and ROUGE-L scores for LaTeX, DSL, Wiki, and Code tasks. For example, compared to Gemini 1.5 Flash, FineEdit-L improves overall BLEU scores by roughly 8\%, with even larger gains observed in specific categories. Notably, FineEdit-XL performs similarly to FineEdit-L, suggesting that increasing the parameter count from 1B to 3B using LLaMA does not yield a significant performance boost.

By leveraging the superior instruction-following capabilities of Qwen2.5-3B-Instruct, our final variant, FineEdit-Pro, further elevates performance. FineEdit-Pro achieves an overall BLEU score of 0.9245, which represents improvements of approximately 11.6\% over Gemini 1.5 Flash, and gains of around 14.7\% and 11.7\% in the DSL and Wiki tasks, respectively. These consistent improvements across multiple data categories underscore the effectiveness of our supervised fine-tuning strategy and highlight the importance of a strong instruction-tuned base model over merely increasing model size.

We also compared our models with Gemini's few-shot prompting approach in real-world scenarios. Although in-context learning (ICL) improves Gemini’s performance in certain cases, such as an 8\% increase in BLEU score on the Wiki dataset for Gemini 2.0 Flash, the overall performance remains inferior to FineEdit-Pro. This superior performance highlights the effectiveness of our high-quality, rigorously validated InstrEditBench dataset in enabling more robust and generalizable solutions for editing tasks.
\begin{table*}[!t]
\centering
\resizebox{0.97\textwidth}{!}{%
\begin{tabular}{ll}
\toprule
\rowcolor{TaskBG}
\textbf{Edit Request 1:} &
  Change the brackets in the code to semicolons. \\[-2pt]
\rowcolor{TaskBG}
\textbf{Original:} &
  \texttt{def test(options): options.data = [] ; def test2(options): options.data = []} \\

\rowcolor{TaskBG}
\textbf{Gemini (\xmark)} &
  \texttt{def test(options): options.data = \textcolor{PrimaryRed}{None} ;
        def test2(options): options.data = \textcolor{PrimaryRed}{None}} \\[-2pt]
\rowcolor{TaskBG}
\textbf{FindEdit Pro (\cmark)} &
  \texttt{def test(options): options.data = \textcolor{PrimaryBlue}{;} ;
        def test2(options): options.data = \textcolor{PrimaryBlue}{;}} \\

\midrule
\rowcolor{TaskBG}
\textbf{Edit Request 2:} &
  Change \texttt{\textbackslash subsection\{Strengths\}} to
  \texttt{\textbackslash subsection*\{Strengths\}}. \\[-2pt]
\rowcolor{TaskBG}
\textbf{Original:} &
  \texttt{\textbackslash subsection\{Strengths\} The topic of responsible AI is ...} \\

\rowcolor{TaskBG}
\textbf{Gemini (\cmark)} &
  \texttt{\textcolor{PrimaryBlue}{\textbackslash subsection*\{Strengths\}} The topic of responsible AI is ...} \\[-2pt]
\rowcolor{TaskBG}
\textbf{FindEdit Pro (\xmark)} &
  \texttt{\textcolor{PrimaryRed}{latex\{\textbackslash subsection\{Strengths\}\}} The topic of responsible AI is ...} \\

\midrule
\rowcolor{TaskBG}
\textbf{Edit Request 3:} &
  Replace ``Falcon'' with ``Captain America''. \\[-2pt]
\rowcolor{TaskBG}
\textbf{Original:} &
  In ``Captain America: Brave New World,'' Sam Wilson, formerly the Falcon, assumes ... \\

\rowcolor{TaskBG}
\textbf{Gemini (\xmark)} &
  In ``Captain America: Brave New World,'' Sam Wilson, formerly
  \textcolor{PrimaryRed}{known as the Falcon}, assumes ... \\[-2pt]
\rowcolor{TaskBG}
\textbf{FindEdit Pro (\cmark)} &
  In ``Captain America: Brave New World,'' Sam Wilson, formerly the
  \textcolor{PrimaryBlue}{Captain America}, assumes ... \\

\midrule
\rowcolor{TaskBG}
\textbf{Edit Request 4:} &
  Add column \texttt{created\_at TIMESTAMP DEFAULT CURRENT\_TIMESTAMP}. \\[-2pt]
\rowcolor{TaskBG}
\textbf{Original:} &
  \texttt{CREATE TABLE worker\_salaries (employee\_id INT, ...)} \\

\rowcolor{TaskBG}
\textbf{Gemini (\xmark)} &
  \textcolor{PrimaryRed}{\texttt{ALTER TABLE community\_gardens ADD COLUMN created\_at TIMESTAMP DEFAULT CURRENT\_TIMESTAMP}} \\[-2pt]
\rowcolor{TaskBG}
\textbf{FindEdit Pro (\cmark)} &
  \texttt{CREATE TABLE community\_gardens (...,
    \textcolor{PrimaryBlue}{created\_at TIMESTAMP DEFAULT CURRENT\_TIMESTAMP})} \\

\bottomrule
\end{tabular}}
\caption{Colour-coded comparison of Gemini and FindEdit Pro responses to four edit requests. (\xmark\ = incorrect) stands for incorrect editing, while (\cmark\ = correct) stands for correct editing.}
\label{tab:gemini_finedit_comparison}
\end{table*}
\subsection{Multi-turn Editing Evaluation}
We also evaluated the extended benchmark on FineEdit in the multi-turn setting. In this extension, each data instance contains multiple editing requests. To simulate real multi-turn scenario, we apply these instructions iteratively: each request is executed on the output produced by the previous one, ensuring that the edits are applied in a cumulative manner. We then assess the final output to determine whether it accurately reflects the cumulative effect of all editing instructions after the full sequence of modifications has been applied. To assess the performance of multi-turn, we randomly sample 100 multi-turn data for each category, and test them on different models. 

Results show that multi-turn editing leads to consistent performance drops across all domains. Specifically, BLEU scores for LaTeX drop from 0.9539 to 0.8461, DSL from 0.9521 to 0.8123, Wiki from 0.8521 to 0.6975, and Code from 0.9538 to 0.9499. These results indicate that multi-turn scenarios are substantially more challenging, especially for Wiki and DSL, while code exhibit strong robustness under multi-turn edits. This indicates that the decline is the accumulation of errors across turns. Each instruction is applied to the output of the previous one, which may already contain small deviations. These deviations can propagate through subsequent steps and lead to compounded errors in the final output.

Despite the degradation in multi-turn settings, FineEdit-Pro achieves an average BLEU of 0.8265, substantially higher than Gemini 1.5 Flash (0.5764) and Gemini 2.0 Flash (0.3899). This further demonstrates the effectiveness of our dataset design and its extensibility to diverse editing scenarios.

\subsection{Qualitative Study}
To qualitatively assess the performance of FindEdit on single-turn editing tasks, we conduct several studies as shown in Table~\ref{tab:gemini_finedit_comparison}. This table illustrates eight examples of how FineEdit-Pro and Gemini respond to diverse editing requests. In several cases, FineEdit-Pro accurately applies the required changes. Specifically, it could correctly add new columns in DSL or adjust environment commands. However, Gemini often restates the instruction without actually implementing the intended modifications.
\begin{table*}[!t]
\centering
\resizebox{0.97\textwidth}{!}{%
\begin{tabular}{ll}
\toprule
\rowcolor{TaskBG}
\textbf{Edit Request 1:} &
  \texttt{Remove the duplicate \textbackslash begin\{abstract\} at the beginning of the abstract environment}. \\[-2pt]
\rowcolor{TaskBG}
\textbf{Original:} &
  \texttt{\textbackslash begin\{abstract\} \textbackslash begin\{abstract\} …} \\

\rowcolor{TaskBG}
\textbf{Gemini (\xmark)} &
  \texttt{\textbackslash begin\{abstract\} … \textbackslash end\{abstract\}
        \textcolor{PrimaryRed}{\textbackslash end\{abstract\}}} \\[-2pt]
\rowcolor{TaskBG}
\textbf{FindEdit Pro (\cmark)} &
  \texttt{\textcolor{PrimaryBlue}{\textbackslash begin\{abstract\}} … 
        \textcolor{PrimaryBlue}{\textbackslash end\{abstract\}}} \\

\midrule
\rowcolor{TaskBG}
\textbf{Edit Request 2:} &
  Convert \texttt{\textbackslash footnote\{...\}} to
  \texttt{\textbackslash footnotemark} + \texttt{\textbackslash footnotetext\{...\}}. \\[-2pt]
\rowcolor{TaskBG}
\textbf{Original:} &
  \texttt{\textbackslash footnote\{Dataset is available at \textbackslash url\{...\}\}} \\

\rowcolor{TaskBG}
\textbf{Gemini (\xmark)} &
  \texttt{\textbackslash footnote\{\textcolor{PrimaryRed}{Dataset is available at \textbackslash url\{...\}}\}} \\[-2pt]
\rowcolor{TaskBG}
\textbf{FindEdit Pro (\cmark)} &
  \texttt{\textcolor{PrimaryBlue}{\textbackslash footnotemark}}
  \quad … \quad
  \texttt{\textcolor{PrimaryBlue}{\textbackslash footnotetext\{Dataset is available at \textbackslash url\{...\}\}}} \\

\midrule
\rowcolor{TaskBG}
\textbf{Edit Request 3:} &
  Remove the 'TODO' from the 'TODO' line. \\[-2pt]
\rowcolor{TaskBG}
\textbf{Original:} &
  \texttt{\textbackslash \{TODO: we introduce distractibility as a new metric for evaluating language models.\}} \\

\rowcolor{TaskBG}
\textbf{Gemini (\cmark)} &
  \texttt{\textcolor{PrimaryBlue}{\textbackslash \{We introduce distractibility as a new metric for evaluating language models.\}}} \\[-2pt]
\rowcolor{TaskBG}
\textbf{FindEdit Pro (\cmark)} &
  \texttt{\textcolor{PrimaryBlue}{\textbackslash \{We introduce distractibility as a new metric for evaluating language models.\}}} \\

\bottomrule
\end{tabular}}
\caption{Color-coded comparison of Gemini and the FineEdit Pro for a multi-turn task with three edit requests. (\xmark\ = incorrect) indicates an unsatisfied edit, while (\cmark\ = correct) indicates a satisfied edit.}
\label{tab:edit_comparison}
\end{table*}
Specifically, both Gemini 1.5 Flash and 2.0 Flash perform well on LaTeX and Wiki tasks, yet they struggle with DSL and Code tasks. For example, as shown in Table~\ref{tab:gemini_finedit_comparison}, FineEdit-Pro correctly identifies the target table and appends a new column named \texttt{created\_at} with the data type \texttt{DEFAULT CURRENT\_TIMESTAMP}. In contrast, Gemini misinterprets the instruction, merely repeating the edit request rather than applying the intended change. These observations highlight the qualitative strengths of our proposed FineEdit approach.

Nonetheless, FineEdit is not without shortcomings. In the LaTeX example depicted in Table~\ref{tab:gemini_finedit_comparison}, Gemini accurately locates the \texttt{\/subsection\{Strengths\}} and updates it as specified. However, although FineEdit-Pro also identifies and modifies the correct location, it generates the correct response twice, which deviates from the direct editing requirement. This discrepancy suggests that FineEdit-Pro, though generally more reliable, can overapply modifications in specific cases.

Overall, these results illustrate FineEdit-Pro’s capacity to handle more complex edits, particularly for DSL and Code, while Gemini often fails to implement them. Nevertheless, occasional issues like duplicate outputs highlight the need for refinement, ensuring FineEdit-Pro consistently adheres to direct editing requirements without introducing redundant content. On the other hand, Gemini occasionally performs better in simpler tasks, such as LaTeX updates.

FineEdit is also generalized well to the multi-turn editing task scenarios. 
Table \ref{tab:edit_comparison} demonstrates an example of how FineEdit-Pro perform more precisely than Gemini.
In this scenario, FineEdit-Pro successfully applies all three requested changes: it removes the duplicate \texttt{\textbackslash begin\{abstract\}}, replaces the inline \texttt{\textbackslash footnote\{\}} with a \texttt{\textbackslash footnotemark} and corresponding \texttt{\textbackslash footnotetext\{\}} pair, and rewrites the commented \texttt{\textbackslash \{TODO...\}} as a finalized, explanatory sentence. 

In contrast, though Gemini removes the duplicate \texttt{\textbackslash begin\{abstract\}}, it add a duplicate 
\texttt{\textbackslash end\{abstract\}} in the end of this content. Additionally, Gemini does not follow the instruction to split the {\textbackslash footnote\{\}} into \texttt{\textbackslash footnotemark} and \texttt{\textbackslash footnotetext\{\}}, instead simply retaining the original inline footnote. These errors indicate that Gemini struggles with compound edits that involve structural modifications across multiple locations.

\subsection{Human Evaluation}
To assess whether \texttt{DiffEval} improves the overall quality of the dataset, we carried out a human evaluation. Because the dataset includes Code and DSL categories that require programming expertise, we recruited three evaluators, each with at least a bachelor’s degree in computer science or a related discipline. We established the following guidelines to ensure rigorous assessment:
(1) {Precise Observation:} Confirm that the updated content exactly corresponds to the segment specified by the edit request. (2) {No Unintended Modifications:} Verify that no other sections have been altered; any unexpected changes result in failure. (3) {Three-Round Procedure:} Two evaluators independently review each item, with a third evaluator resolving any discrepancies.

We examined 100 items per category and found that data processed through our DiffEval pipeline exhibited noticeably enhanced accuracy, as shown in Table~\ref{tab:human_eval}. The Wiki and Code datasets, in particular, demonstrated the most reliable outcomes, with edited content precisely matching the requested modifications. Notably, the DSL dataset experienced the greatest improvement, with quality increasing by over 24\% compared to data that did not meet DiffEval’s standards.

\begin{table}[!tbp]
\centering
\renewcommand\arraystretch{1.1}
\resizebox{0.45\textwidth}{!}{
\begin{tabular}{lcccc}
\toprule
\textbf{Threshold} & \textbf{Wiki} & \textbf{LaTeX} & \textbf{DSL} & \textbf{Code} \\
\midrule
\textbf{G-score $\geq$ 9} & 97\% & 93\% & 90\% & 97\% \\
\textbf{G-score $<$ 9}    & 87\% & 89\% & 66\% & 83\% \\
\bottomrule
\end{tabular}
}
\caption{Annotation accuracy across content types under different G-score thresholds.}
\label{tab:human_eval}
\vspace{-5mm}
\end{table}

\subsection{Ablation Study on DiffEval Components}

To better understand the contribution of each component in the DiffEval pipeline, we conducted two ablation experiments. Manual annotation followed the protocol described in Section 4.6.

\paragraph{\texttt{Git diff} effectiveness.}
We evaluated a reduced pipeline in which G-Eval judged the alignment between the instruction and the edited text without access to \texttt{git diff}. From its output, we randomly sampled \texttt{100} examples whose \texttt{G-Score} was at least nine and annotated them. For comparison, we annotated another \texttt{100} examples produced by the full DiffEval pipeline, where G-Eval received the \texttt{git diff} instead of the full edited text.

Including \texttt{git diff} raised the accuracy from 0.85 to 0.94. These results indicate that \texttt{git diff} contributes important structural information for identifying precise alignment between the instruction and the edit.

\paragraph{G-score threshold selection.}
We also examined the effect of the G-score threshold $\alpha$ used in filtering. Setting $\alpha = 8$ results in many examples where the core instruction is followed, but this will sometimes introduce unintended formatting changes. One common issue is the insertion of extra spaces throughout the text. For instance, in a deletion instruction targeting a historical phrase, the phrase was correctly removed, but the resulting diff introduced multiple superfluous spaces across the paragraph. This violates the requirement to preserve all formatting outside the instructed change. Such formatting issues were significantly reduced when the threshold was increased to $\alpha = 9$.

\section{Conclusion}

We introduce InstrEditBench, a benchmark of over 30k editing tasks spanning Wiki, LaTeX, code, and DSL, aimed at precise instruction-based text editing. To ensure supervision quality, we develop DiffEval, an automated pipeline combining structural and semantic filters. We further validate our benchmark with FineEdit, a model fine-tuned on InstrEditBench, achieving up to 10\% gains over leading models. Designed for both single-turn and multi-turn editing, our modular benchmark and pipeline enable broad applicability.  

\section{Limitations}

\noindent \textbf{Limited Deployment Scope.} Due to cost and hardware constraints, our evaluations were limited to large proprietary LLMs (e.g., Gemini), rather than large open-source models.

\noindent \textbf{Controlled Context Evaluation.} Our benchmark focuses on controlled evaluation contexts, where it does not yet encompass long-context chain-of-thought scenarios, as smaller LLMs are confined by limited context windows, even though such techniques could be effective in proprietary models.

\section{Acknowledgement}
This work was supported by the National Science Foundation under Grant No. CCF-2403747. The opinions, findings, and conclusions or recommendations expressed herein are those of the authors and do not necessarily reflect the views of the National Science Foundation. Meanwhile, we sincerely thank the reviewers for their valuable feedback, which greatly contributed to improving this work.

\bibliography{anthology,custom}

@misc{lei2025infantagentnextmultimodalgeneralistagent,
      title={InfantAgent-Next: A Multimodal Generalist Agent for Automated Computer Interaction}, 
      author={Bin Lei and Weitai Kang and Zijian Zhang and Winson Chen and Xi Xie and Shan Zuo and Mimi Xie and Ali Payani and Mingyi Hong and Yan Yan and Caiwen Ding},
      year={2025},
      eprint={2505.10887},
      archivePrefix={arXiv},
      primaryClass={cs.AI},
      url={https://arxiv.org/abs/2505.10887}, 
}

@misc{liang2025autoranweaktostrongjailbreakinglarge,
      title={AutoRAN: Weak-to-Strong Jailbreaking of Large Reasoning Models}, 
      author={Jiacheng Liang and Tanqiu Jiang and Yuhui Wang and Rongyi Zhu and Fenglong Ma and Ting Wang},
      year={2025},
      eprint={2505.10846},
      archivePrefix={arXiv},
      primaryClass={cs.LG},
      url={https://arxiv.org/abs/2505.10846}, 
}

@INPROCEEDINGS{regulogpt,
  author={Wu, Xidong and Jo, Sumin and Zeng, Yiming and Das, Arun and Zhang, Ting-He and Patel, Parth and Wei, Yuanjing and Li, Lei and Gao, Shou-Jiang and Zhang, Jianqiu and Pratt, Dexter and Chiu, Yu-Chiao and Huang, Yufei},
  booktitle={2024 IEEE EMBS International Conference on Biomedical and Health Informatics (BHI)}, 
  title={ReguloGPT: Harnessing GPT for End-to-End Knowledge Graph Construction of Molecular Regulatory Pathways}, 
  year={2024},
  volume={},
  number={},
  pages={1-8},
  doi={10.1109/BHI62660.2024.10913581}}

@inproceedings{li-etal-2025-learning-committee,
    title = "Learning from Committee: Reasoning Distillation from a Mixture of Teachers with Peer-Review",
    author = "Li, Zhuochun  and
      Ji, Yuelyu  and
      Meng, Rui  and
      He, Daqing",
    editor = "Che, Wanxiang  and
      Nabende, Joyce  and
      Shutova, Ekaterina  and
      Pilehvar, Mohammad Taher",
    booktitle = "Findings of the Association for Computational Linguistics: ACL 2025",
    month = jul,
    year = "2025",
    address = "Vienna, Austria",
    publisher = "Association for Computational Linguistics",
    url = "https://aclanthology.org/2025.findings-acl.217/",
    doi = "10.18653/v1/2025.findings-acl.217",
    pages = "4190--4205",
    ISBN = "979-8-89176-256-5",
}

@article{castillo2022chat,
  title={Chat GPT: a promising tool for academic editing},
  author={Castillo-Gonz{\'a}lez, William and Lepez, Carlos Oscar and Bonardi, Mabel Cecilia},
  journal={Data and Metadata},
  volume={1},
  pages={23--23},
  year={2022}
}

@misc{google2024gemini,
  author       = {Google DeepMind},
  title        = {Google Gemini AI Update - December 2024},
  year         = {2024},
  url          = {https://blog.google/technology/google-deepmind/google-gemini-ai-update-december-2024/},
}

@misc{meta2024llama3_2,
  author       = {{Meta AI}},
  title        = {Llama 3.2: Advancing Vision and Edge AI for Mobile Devices},
  year         = {2024},
  url          = {https://ai.meta.com/blog/llama-3-2-connect-2024-vision-edge-mobile-devices/},
  note         = {Accessed: 2025-01-06}
}

@misc{raheja2023coedit,
  title     = {CoEdIT: Text Editing by Task-Specific Instruction Tuning},
  author    = {Vipul Raheja and Dhruv Kumar and Ryan Koo and Dongyeop Kang},
  howpublished = {arXiv preprint arXiv:2305.09857},
  year      = {2023}
}

@article{dilhara2024unprecedented,
  title={Unprecedented code change automation: The fusion of llms and transformation by example},
  author={Dilhara, Malinda and Bellur, Abhiram and Bryksin, Timofey and Dig, Danny},
  journal={Proceedings of the ACM on Software Engineering},
  volume={1},
  number={FSE},
  pages={631--653},
  year={2024},
  publisher={ACM New York, NY, USA}
}

@article{raffel2020exploring,
  title={Exploring the limits of transfer learning with a unified text-to-text transformer},
  author={Raffel, Colin and others},
  journal={Journal of Machine Learning Research},
  volume={21},
  number={140},
  pages={1--67},
  year={2020}
}

@inproceedings{pezeshkpour2023measuring,
  title={Measuring and modifying factual knowledge in large language models},
  author={Pezeshkpour, Pouya},
  booktitle={2023 international conference on machine learning and applications (ICMLA)},
  pages={831--838},
  year={2023},
  organization={IEEE}
}

@article{bout2023efficient,
  title={Efficient grammatical error correction via multi-task training and optimized training schedule},
  author={Bout, Andrey and Podolskiy, Alexander and Nikolenko, Sergey and Piontkovskaya, Irina},
  journal={arXiv preprint arXiv:2311.11813},
  year={2023}
}

@inproceedings{bryant2019bea,
    title = "The {BEA}-2019 Shared Task on Grammatical Error Correction",
    author = "Bryant, Christopher  and
      Felice, Mariano  and
      Andersen, {\O}istein E.  and
      Briscoe, Ted",
    booktitle = "Proceedings of the Fourteenth Workshop on Innovative Use of NLP for Building Educational Applications",
    month = aug,
    year = "2019",
    address = "Florence, Italy",
    publisher = "Association for Computational Linguistics",
    url = "https://aclanthology.org/W19-4406/",
    doi = "10.18653/v1/W19-4406",
    pages = "52--75",
}

@inproceedings{stahlberg2021synthetic,
    title = "Synthetic Data Generation for Grammatical Error Correction with Tagged Corruption Models",
    author = "Stahlberg, Felix and Kumar, Shankar",
    booktitle = "Proceedings of the 16th Workshop on Innovative Use of NLP for Building Educational Applications",
    month = apr,
    year = "2021",
    address = "Online",
    publisher = "Association for Computational Linguistics",
    url = "https://www.aclweb.org/anthology/2021.bea-1.4",
    pages = "37--47",
}

@inproceedings{katinskaia2023grammatical,
  title={Grammatical error correction for sentence-level assessment in language learning},
  author={Katinskaia, Anisia and Yangarber, Roman},
  booktitle={Workshop on Innovative Use of NLP for Building Educational Applications},
  pages={488--502},
  year={2023},
  organization={The Association for Computational Linguistics}
}

@article{sun2023teaching,
  title={Teaching the pre-trained model to generate simple texts for text simplification},
  author={Sun, Renliang and Xu, Wei and Wan, Xiaojun},
  journal={arXiv preprint arXiv:2305.12463},
  year={2023}
}

@article{palivela2021optimization,
  title={Optimization of paraphrase generation and identification using language models in natural language processing},
  author={Palivela, Hemant},
  journal={International Journal of Information Management Data Insights},
  volume={1},
  number={2},
  pages={100025},
  year={2021},
  publisher={Elsevier}
}

@inproceedings{luo2023prompt,
  title={Prompt-based editing for text style transfer},
  author={Luo, Guoqing and Han, Yu and Mou, Lili and Firdaus, Mauajama},
  booktitle={Findings of the Association for Computational Linguistics: EMNLP 2023},
  pages={5740--5750},
  year={2023}
}

@article{husain2019codesearchnet,
  title={{CodeSearchNet} challenge: Evaluating the state of semantic code search},
  author={Husain, Hamel and Wu, Ho-Hsiang and Gazit, Tiferet and Allamanis, Miltiadis and Brockschmidt, Marc},
  journal={arXiv preprint arXiv:1909.09436},
  year={2019}
}

@misc{merity2016pointer,
      title={Pointer Sentinel Mixture Models},
      author={Stephen Merity and Caiming Xiong and James Bradbury and Richard Socher},
      year={2016},
      eprint={1609.07843},
      archivePrefix={arXiv},
      primaryClass={cs.CL}
}

@misc{latex2poster,
  title = {LaTeX2Poster Dataset},
  author = {{Latex2Poster}},
  year = {2024},
  howpublished = {Hugging Face},
  url = {https://huggingface.co/datasets/jd445/latex2poster},
}

@misc{b-mc2_2023_sql-create-context,
  title   = {sql-create-context Dataset},
  author  = {b-mc2}, 
  year    = {2023},
  url     = {https://huggingface.co/datasets/b-mc2/sql-create-context},
}

@article{shen2017style,
  title={Style transfer from non-parallel text by cross-alignment},
  author={Shen, Tianxiao and Lei, Tao and Barzilay, Regina and Jaakkola, Tommi},
  journal={Advances in neural information processing systems},
  volume={30},
  year={2017}
}

@misc{lian2023mistralorca1,
  title = {MistralOrca: Mistral-7B Model Instruct-tuned on Filtered OpenOrcaV1 GPT-4 Dataset},
  author = {Wing Lian and Bleys Goodson and Guan Wang and Eugene Pentland and Austin Cook and Chanvichet Vong and "Teknium"},
  year = {2023},
  publisher = {HuggingFace},
  journal = {HuggingFace repository},
  howpublished = {https://huggingface.co/Open-Orca/Mistral-7B-OpenOrca},
}

@misc{mukherjee2023orca,
      title={Orca: Progressive Learning from Complex Explanation Traces of GPT-4}, 
      author={Subhabrata Mukherjee and Arindam Mitra and Ganesh Jawahar and Sahaj Agarwal and Hamid Palangi and Ahmed Awadallah},
      year={2023},
      eprint={2306.02707},
      archivePrefix={arXiv},
      primaryClass={cs.CL}
}

@misc{longpre2023flan,
      title={The Flan Collection: Designing Data and Methods for Effective Instruction Tuning}, 
      author={Shayne Longpre and Le Hou and Tu Vu and Albert Webson and Hyung Won Chung and Yi Tay and Denny Zhou and Quoc V. Le and Barret Zoph and Jason Wei and Adam Roberts},
      year={2023},
      eprint={2301.13688},
      archivePrefix={arXiv},
      primaryClass={cs.AI}
}

@misc{cassano2024editevaluatingabilitylarge,
      title={Can It Edit? Evaluating the Ability of Large Language Models to Follow Code Editing Instructions}, 
      author={Federico Cassano and Luisa Li and Akul Sethi and Noah Shinn and Abby Brennan-Jones and Jacob Ginesin and Edward Berman and George Chakhnashvili and Anton Lozhkov and Carolyn Jane Anderson and Arjun Guha},
      year={2024},
      eprint={2312.12450},
      archivePrefix={arXiv},
      primaryClass={cs.SE},
      url={https://arxiv.org/abs/2312.12450}, 
}

@misc{wang2025understandingcharacteristicscodegeneration,
      title={Towards Understanding the Characteristics of Code Generation Errors Made by Large Language Models}, 
      author={Zhijie Wang and Zijie Zhou and Da Song and Yuheng Huang and Shengmai Chen and Lei Ma and Tianyi Zhang},
      year={2025},
      eprint={2406.08731},
      archivePrefix={arXiv},
      primaryClass={cs.SE},
      url={https://arxiv.org/abs/2406.08731}, 
}

@inproceedings{nakamachi2020text,
  title={Text simplification with reinforcement learning using supervised rewards on grammaticality, meaning preservation, and simplicity},
  author={Nakamachi, Akifumi and Kajiwara, Tomoyuki and Arase, Yuki},
  booktitle={Proceedings of the 1st Conference of the Asia-Pacific Chapter of the Association for Computational Linguistics and the 10th International Joint Conference on Natural Language Processing: Student Research Workshop},
  pages={153--159},
  year={2020}
}

@misc{hive,
  author = {hive},
  title = {Apache Hive},
  year = {2024},
  howpublished = {GitHub Repository},
  url = {https://github.com/apache/hive},
}

@misc{cassandra,
  author = {cassandra},
  title = {Apache Cassandra},
  year = {2024},
  howpublished = {GitHub Repository},
  url = {https://github.com/apache/cassandra},
}

@misc{chinookDatabase,
  author = {Lerocha},
  title = {Chinook Database},
  year = {2024},
  howpublished = {GitHub Repository},
  url = {https://github.com/lerocha/chinook-database},
}

@article{dong2024survey,
  author    = {Qingxiu Dong and Liangming Pan and Duyu Tang and Ming Gong and Nan Duan and Heyan Huang and Xiaoyan Zhu},
  title     = {A Survey on In-context Learning},
  journal   = {arXiv preprint arXiv:2301.00234},
  year      = {2024},
  url       = {https://arxiv.org/abs/2301.00234}
}

@misc{gitdiff,
  
  title     = {Git Diff: A Tool for Comparing Changes},
  year      = {2024},
  howpublished = {Git Documentation},
  url       = {https://git-scm.com/docs/git-diff},
}

@inproceedings{liu2023geval,
  title     = {G-Eval: NLG Evaluation using GPT-4 with Better Human Alignment},
  author    = {Yang Liu and Dan Iter and Yichong Xu and Shuohang Wang and Ruochen Xu and Chenguang Zhu},
  booktitle = {Proceedings of the 2023 Conference on Empirical Methods in Natural Language Processing},
  pages     = {2511--2522},
  year      = {2023},
  url       = {https://aclanthology.org/2023.emnlp-main.153/},
  doi       = {10.18653/v1/2023.emnlp-main.153}
}

@article{suri2024docedit,
  title={DocEdit-v2: Document Structure Editing Via Multimodal LLM Grounding},
  author={Suri, Manan and Mathur, Puneet and Dernoncourt, Franck and Jain, Rajiv and Morariu, Vlad I and Sawhney, Ramit and Nakov, Preslav and Manocha, Dinesh},
  journal={arXiv preprint arXiv:2410.16472},
  year={2024}
}

@misc{fan2024exploringcapabilitiesllmscode,
      title={Exploring the Capabilities of LLMs for Code Change Related Tasks}, 
      author={Lishui Fan and Jiakun Liu and Zhongxin Liu and David Lo and Xin Xia and Shanping Li},
      year={2024},
      eprint={2407.02824},
      archivePrefix={arXiv},
      primaryClass={cs.SE},
      url={https://arxiv.org/abs/2407.02824}, 
}

@article{hu2021lora,
  title={Lora: Low-rank adaptation of large language models},
  author={Hu, Edward J and Shen, Yelong and Wallis, Phillip and Allen-Zhu, Zeyuan and Li, Yuanzhi and Wang, Shean and Wang, Lu and Chen, Weizhu},
  journal={arXiv preprint arXiv:2106.09685},
  year={2021}
}

@article{wang2022self,
  title={Self-instruct: Aligning language models with self-generated instructions},
  author={Wang, Yizhong and Kordi, Yeganeh and Mishra, Swaroop and Liu, Alisa and Smith, Noah A and Khashabi, Daniel and Hajishirzi, Hannaneh},
  journal={arXiv preprint arXiv:2212.10560},
  year={2022}
}

@misc{alpaca,
  author = {Rohan Taori and Ishaan Gulrajani and Tianyi Zhang and Yann Dubois and Xuechen Li and Carlos Guestrin and Percy Liang and Tatsunori B. Hashimoto },
  title = {Stanford Alpaca: An Instruction-following LLaMA model},
  year = {2023},
  publisher = {GitHub},
  journal = {GitHub repository},
  howpublished = {\url{https://github.com/tatsu-lab/stanford_alpaca}},
}

@misc{wikipedia_featured_articles,
  author = {{Wikipedia}},
  title = {Wikipedia: Featured articles},
  howpublished = {\url{https://en.wikipedia.org/wiki/Wikipedia:Featured_articles}},
  note = {Accessed: 2025-02-14},
  year = {n.d.}
}

@misc{wikipedia_good_articles,
  author = {{Wikipedia}},
  title = {Wikipedia: Good articles},
  howpublished = {\url{https://en.wikipedia.org/wiki/Wikipedia:Good_articles}},
  note = {Accessed: 2025-02-14},
  year = {n.d.}
}

@article{madaan2023self,
  title={Self-refine: Iterative refinement with self-feedback},
  author={Madaan, Aman and Tandon, Niket and Gupta, Prakhar and Hallinan, Skyler and Gao, Luyu and Wiegreffe, Sarah and Alon, Uri and Dziri, Nouha and Prabhumoye, Shrimai and Yang, Yiming and others},
  journal={Advances in Neural Information Processing Systems},
  volume={36},
  pages={46534--46594},
  year={2023}
}

@article{schick2022peer,
  title={Peer: A collaborative language model},
  author={Schick, Timo and Dwivedi-Yu, Jane and Jiang, Zhengbao and Petroni, Fabio and Lewis, Patrick and Izacard, Gautier and You, Qingfei and Nalmpantis, Christoforos and Grave, Edouard and Riedel, Sebastian},
  journal={arXiv preprint arXiv:2208.11663},
  year={2022}
}

@article{achiam2023gpt,
  title={Gpt-4 technical report},
  author={Achiam, Josh and Adler, Steven and Agarwal, Sandhini and Ahmad, Lama and Akkaya, Ilge and Aleman, Florencia Leoni and Almeida, Diogo and Altenschmidt, Janko and Altman, Sam and Anadkat, Shyamal and others},
  journal={arXiv preprint arXiv:2303.08774},
  year={2023}
}

@article{celikyilmaz2020evaluation,
  title={Evaluation of text generation: A survey},
  author={Celikyilmaz, Asli and Clark, Elizabeth and Gao, Jianfeng},
  journal={arXiv preprint arXiv:2006.14799},
  year={2020}
}

\clearpage

\appendix

\section{Additional Implementation Details}
\label{sec:appx_implementation_detail}

For existing models, we strictly adhere to configurations from their original papers. To manage fixed maximum token lengths \(L\), if the combined \(T_{\text{orig}}\) and \(I_{\text{edit}}\) exceed \(L\), we partition \(T_{\text{orig}}\) into chunks of size \(\leq L\), process each chunk independently with the same edit instruction, and concatenate the outputs to form the complete edited text. We fine-tune models using Low-Rank Adaptation (LoRA) \cite{hu2021lora} with \(r=8\), \(\alpha=32\), and a dropout rate of 0.05, employing the AdamW optimizer with a learning rate of \(2 \times 10^{-5}\), training for 2 epochs, an effective batch size of 1, and 4 gradient accumulation steps.

\noindent \textbf{Chunking long context:}
Many large language models impose a fixed maximum token length \( L \) on their input (and sometimes output) sequences. Consequently, if the combination of \( T_{\text{orig}} \) and \( I_{\text{edit}} \) exceeds this limit, we divide the \( T_{\text{orig}} \) into smaller chunks of size \( \leq L \). Each chunk is then processed independently—paired with the same edit request and later concatenated to form the complete edited text. This approach ensures that every chunk fits within the model’s token budget, preventing overflow and reducing memory usage while preserving the overall structured editing behavior.

\noindent \textbf{Fine-Tuning Configuration:} We use a LoRA rank of \(\texttt{r = 8}\) and LoRA alpha \(\alpha = 32\), following the original LoRA paper~\cite{hu2021lora}. 
This combination (\(\texttt{r = 8}\), \(\alpha = 32\)) and a learning rate of \(2 \times 10^{-5}\) are widely used in practice, including in the default settings of the HuggingFace PEFT library. 
It produces a scaling factor of \(\alpha / r = 4\), which balances training stability and memory efficiency, enabling the model to learn meaningful updates without destabilizing training.
We set \(\texttt{lora\_dropout = 0.05}\), a typical value that helps regularize LoRA updates and reduce overfitting.
Training and Generation Settings are as follows:
\begin{itemize} 
    \item \textbf{Epochs:} 2 epochs, which is generally sufficient for convergence in our editing task.
    \item \textbf{Gradient Accumulation Steps:} 4 (necessary due to a small batch size of 1 and GPU constraints).
    \item \textbf{Max Chunk Tokens:} 2048.
    \item \textbf{Max Length:} 4096.
    \item \textbf{Generation Settings:} \(\texttt{temperature = 0.2}\), \(\texttt{top-p = 0.95}\).
\end{itemize}
The token constraints ensure no exceeding of the model's context window and maintain consistent training across models. These parameters reduce randomness while keeping the generated text relevant to the task.

\noindent \textbf{Decoding and Inference:}  
During generation, we set the temperature to 0.2 and used top-p sampling with a probability of 0.95, then merging outputs from all chunks to produce the final edited text. The temperature and top-p settings follow previous editing task studies \cite{cassano2024editevaluatingabilitylarge} to ensure minimal changes rather than creative expansions as our editing tasks require precise.
\section{Data Example}
Table~\ref{tab:edit-intentions} presents representative examples from our benchmark, covering four distinct data categories—WikiText, LaTeX, Code, and Database DSL. Each example includes the original content, the user-issued edit request, the resulting edited content, the line-level difference, and the associated G-score indicating edit difficulty.

We make a concrete instance using data in the LaTeX category in Table~\ref{tab:edit-intentions}. If the edit request is to ``Remove the duplicate \texttt{\textbackslash{}begin\{abstract\}} at the beginning of the abstract environment," the diff output might display on Line~1:
\begin{verbatim}
\begin{abstract}[-\begin{abstract}-]
\end{verbatim}
This indicates that the duplicate has been successfully removed.

\section{Dataset Generation Prompts}
\label{sec:dataset_generation_prompts}

We use the following prompts for dataset generation on each domain.
\begin{tcolorbox}
\small
\texttt{
user\_prompt = r'''Task: Generate one precise editing request for the given LaTeX code, focusing exclusively on one detailed LaTeX-specific aspect.\\
\ \ \ \ \ 1. Analyze LaTeX Components: Examine the LaTeX code thoroughly, identifying elements such as commands, environments, packages, mathematical expressions, figures, tables, references, labels, and syntax structures.\\
\ \ \ \ \ 2. Target a Single LaTeX Issue: The editing request must address only one specific LaTeX-related issue such as commands, environments, packages, mathematical expressions, figures, tables, references, labels, and syntax structures.\\
\ \ \ \ \ 3. Clearly define the exact edit needed. The action should be definitive and unambiguous, avoiding any form of suggestion, optional language, or choices. Do not include reasons for the edit or any additional information beyond the request.\\
\ \ \ \ \ 4. Do not include reasons for the edit or any additional information beyond the edit request. The request should be a direct instruction.\\
\ \ \ \ \ The request examples are:\\
\ \ \ \ \ [Example 1]\\
\ \ \ \ \ <Edit Request>\\
\ \ \ \ \ Replace the \textbackslash begin\{equation\} ... \textbackslash end\{equation\} environment with a \textbackslash [ ...\textbackslash ] display math environment to present the equation.\\
\ \ \ \ \ </Edit Request>\\
\ \ \ \ \ [Example 2]\\
\ \ \ \ \ <Edit Request>\\
\ \ \ \ \ Remove the \textbackslash centering command inside the figure environment and insert \textbackslash centering immediately after \textbackslash begin\{figure\}.\\
\ \ \ \ \ </Edit Request>\\
\ \ \ \ \ [Example 3]\\
\ \ \ \ \ <Edit Request>\\
\ \ \ \ \ Change the citation command \textbackslash cite\{einstein\} to \textbackslash parencite\{einstein\} to display the citation in parentheses.\\
\ \ \ \ \ </Edit Request>\\
\ \ \ \ \ [Example 4]\\
\ \ \ \ \ <Edit Request>\\
\ \ \ \ \ Change the column specification in the tabular environment from \{l l l\} to \{l c r\} to adjust the alignment of the data columns.\\
\ \ \ \ \ </Edit Request>\\
\ \ \ \ \ [Example 5]\\
\ \ \ \ \ <Edit Request>\\
\ \ \ \ \ Replace the placeholder ??? in the reference text with \textbackslash ref\{sec:relwork\} to properly reference the “Related Work” section.\\
\ \ \ \ \ </Edit Request>\\
\ \ \ \ \ [Example 6]\\
\ \ \ \ \ <Edit Request>\\
\ \ \ \ \ Rename the macro \textbackslash vect to \textbackslash vecbold in both its definition and throughout the document.\\
\ \ \ \ \ </Edit Request>\\
}
\end{tcolorbox}
\begin{tcolorbox}
\small
\texttt{
\ \ \ \ \ [Example 7]\\
\ \ \ \ \ <Edit Request>\\
\ \ \ \ \ Add the optional width argument to \textbackslash includegraphics\{example-image\} as \textbackslash includegraphics[width=0.5\textbackslash textwidth] \\
\ \ \ \ \ \{example-image\} to scale the image.\\
\ \ \ \ \ </Edit Request>\\
\ \ \ \ \ [Example 8]\\
\ \ \ \ \ <Edit Request>\\
\ \ \ \ \ Remove the \textbackslash usepackage\{epsfig\} line and replace it with \textbackslash usepackage\{graphicx\} to handle graphics\\
\ \ \ \ \ </Edit Request>\\
\\
\ \ \ \ \ I will give you the content and then the editing request.\\ 
\ \ \ \ \ Please Edit the content based on the editing request. \\
\ \ \ \ \ While Editing, don't add other words like\\
\ \ \ \ \ modified or something. Just Edit directly. \\
\\
\ \ \ \ \ Content: \{original\_context\} \\
\ \ \ \ \ Editing Request: \{edit\_request\} \\
\ \ \ \ \ Please return the complete content after editing. \\
\ \ \ \ \ Don't skip the empty line and keep the original\\
\ \ \ \ \ apart from the editing part.
}
\end{tcolorbox}

We use the following prompts for G-Eval.
\begin{tcolorbox}
\small
\begin{verbatim}
Understanding Content Differences:
Changes between the original and edited texts 
are categorized and formatted as follows:
Replace (replace): [original_text -> 
modified_text]
Delete (delete): [-original_text-]
Insert (insert): [+modified_text+]
Equal (equal): unchanged_text

Instructions:
1) Read Carefully: Examine the original 
contents and the edited parts (shown as the 
formatted diff) thoroughly.
2) Identify and Evaluate: Using the above diff 
formatting rules, determine if the 
modifications are both correct and complete 
throughout the entire original content.
3) Assign a Coherence Score: Based on the 
Evaluation Criteria, rate the coherence of 
the modifications on a scale of 1 to 10, 
where 1 is the lowest and 10 is the highest.

Final Output: Only provide the numeric 
coherence score.
The given contents are: 1) the entire original 
content, 2) the edit request, and 3) a 
formatted diff that shows how the edited 
content differs from the original.
\end{verbatim}
\end{tcolorbox}

\begin{table*}[h!]
\vspace{-120mm}
\centering
\resizebox{\textwidth}{!}{ 
\begin{tabular}{@{}p{2.5cm}p{5.5cm}p{5.5cm}p{5.5cm}p{6.0cm}p{1.0cm}@{}}
\toprule
\textbf{Data Category} & \textbf{Orignal Content} & \textbf{Edit Request} & \textbf{Edited Content} & \textbf{Difference} & \textbf{G-score}\\ \midrule
\textbf{WikiText} & ...As with previous <unk> Chronicles games, Valkyria Chronicles III is a tactical role @-@ playing game where players take control of a military unit... & Replace ``\textbackslash{}<unk>\textbackslash{}'' with ``Valkyria'' where it appears in the text.
 & ...As with previous Valkyria Chronicles games, Valkyria Chronicles III is a tactical role @-@ playing game where players take control of a military unit... & Line 2 differs: Differences: ...As with previous \textcolor{red}{[<un -> Val]k[> -> yria]} Chronicles games, Valkyria Chronicles III
is a tactical role @-@ playing game
where players take control of a mili-
tary unit... & 9 \\ \midrule 
\textbf{LaTex} & 
\textbackslash{}begin\{abstract\}\textbackslash{}n\textbackslash{}begin\{abstract\}\textbackslash {}n  \%\textbackslash{}mika\{\}, \textbackslash{}guandao\{\}, \textbackslash{}leo\{\}\textbackslash{}n  \textbackslash{}vspace\{-0.2cm\}\textbackslash{}n  Neural radiance fields (NeRF) rely on volume rendering to...
 & Remove the duplicate \textbackslash{}begin\{abstract\} at the beginning of the abstract environment. & 
\textbackslash{}begin\{abstract\}\textbackslash{}n  \%\textbackslash{}mika\{\}, \textbackslash{}guandao\{\},\textbackslash{}leo\{\}\textbackslash{}n  \textbackslash{}vspace\{-0.2cm\}\textbackslash{}n  Neural radiance fields (NeRF) rely on volume rendering to... & Line 1 differs: Differences: \textbackslash{}begin\{abstract\}\textcolor{red}{[- \textbackslash{}begin\{abstract\}-]} & 9
 \\ \midrule
\textbf{Code} & {...def yield\_nanopub(assertions, annotations, line\_num):\textbackslash{}n """Yield nanopub object""" if not assertions:...}
& Change the function definition from:

def yield\_nanopub(assertions, annotations, line\_num)

to include type annotations as:

def yield\_nanopub(assertions: list, annotations: dict, line\_num: int) -> dict
&...def yield\_nanopub(assertions: list, annotations: dict, line\_num: int) -> dict:
 """Yield nanopub object"""
if not assertions:... & Line 1 differs:
Differences: def yield\_nanopub({assertions\textcolor{red}{[+: list+]}, annotations\textcolor{red}{[+: dict+]}, line\_num\textcolor{red}{[+: int+]})\textcolor{red}{[+ -> dict+]}:} & 10
 \\ \midrule

\textbf{Database DSL} & 
...CREATE TABLE DB\_PRIVS\textbackslash{}n
(\textbackslash{}n
DB\_GRANT\_ID NUMBER NOT NULL,\textbackslash{}n
CREATE\_TIME NUMBER (10) NOT NULL,\textbackslash{}n
DB\_ID NUMBER NULL,\textbackslash{}n
)...
 & Rename the column \texttt{"CREATE\_TIME"} in the \texttt{DB\_PRIVS} table to \texttt{"CREATION\_TIMESTAMP"} & ...CREATE TABLE DB\_PRIVS\textbackslash{}n
(\textbackslash{}n
DB\_GRANT\_ID NUMBER NOT NULL,\textbackslash{}n
CREATION\_TIMESTAMP NUMBER (10) NOT NULL,\textbackslash{}n
DB\_ID NUMBER NULL,\textbackslash{}n
)... & Line 4 differs: Differences: CREATE\textcolor{red}{[E \texttt{->}ION]}\_TIME\textcolor{red}{[+STAMP+]} NUMBER (10) NOT NULL,
 & 9 \\ \bottomrule
\end{tabular}
}
\caption{Data examples of different data categories with all attributes (content, edit request, edited content, difference, and G-score).}
\label{tab:edit-intentions}
\end{table*}

\end{document}